\documentclass{article} 
\usepackage{iclr2026_conference,times}

\usepackage{amsmath,amsfonts,bm}

\def\eqref#1{equation~\ref{#1}}

\def\1{\bm{1}}

\DeclareMathAlphabet{\mathsfit}{\encodingdefault}{\sfdefault}{m}{sl}
\SetMathAlphabet{\mathsfit}{bold}{\encodingdefault}{\sfdefault}{bx}{n}

\usepackage[hidelinks]{hyperref}
\usepackage{url}

\usepackage{amsmath} 
\usepackage{amssymb}  
\usepackage{microtype} 
\usepackage{url}
\usepackage{xcolor}
\usepackage{graphicx}
\usepackage{todonotes}
\usepackage[most]{tcolorbox}
\usepackage[linesnumbered,ruled,vlined]{algorithm2e}
\usepackage[switch]{lineno}
\usepackage{subcaption}
\usepackage{svg}
\usepackage{amsmath}
\usepackage{amssymb}
\usepackage{bm}
\usepackage{csquotes}
\usepackage{balance}

\title{Evaluating VLMs' Spatial Reasoning Over Robot Motion: A Step Towards Robot Planning with Motion Preferences}

\author{
  Wenxi Wu\textsuperscript{1} \quad
  Jingjing Zhang\textsuperscript{2} \quad
  Martim Brand\~ao\textsuperscript{1}  \\[0.2cm] 
  \textsuperscript{1}King's College London 
  \quad
  \textsuperscript{2}University College London
}

\iclrfinalcopy 
\begin{document}

\maketitle

\begin{abstract}
Understanding user instructions and object spatial relations in surrounding environments is crucial for intelligent robot systems to assist humans in various tasks. The natural language and spatial reasoning capabilities of Vision-Language Models (VLMs) have the potential to enhance the generalization of robot planners on new tasks, objects, and motion specifications. 
While foundation models have been applied to task planning, it is still unclear the degree to which they have the capability of spatial reasoning required to enforce user preferences or constraints on motion, such as desired distances from objects, topological properties, or motion style preferences.
In this paper, we evaluate the capability of four state-of-the-art VLMs at spatial reasoning over robot motion, using four different querying methods. Our results show that, with the highest-performing querying method, Qwen2.5-VL achieves 71.4\% accuracy zero-shot and 75\% on a smaller model after fine-tuning, and GPT-4o leads to lower performance. We evaluate two types of motion preferences (object-proximity and path-style), and we also analyze the trade-off between accuracy and computation cost in number of tokens. This work shows some promise in the potential of VLM integration with robot motion planning pipelines.
\end{abstract}

\section{Introduction}

Spatial reasoning can enable robots to understand the relationships among objects in the environment, which is essential for robots to interact with the physical world. With the abundant semantic knowledge from internet-scale data, foundation models' ability to mimic natural language  understanding and visual reasoning has to potential to provide an intuitive interface for users to give instructions to robots. 

More specifically, our assumption in this paper is that Visual Language Models (VLMs) can enable users to express preferences over robot motion. For instance, preferences over motion could be about whether users would like a long or wavy path, or a smooth or rough path, or a path that passes close or far away from an object (e.g. ``move the can while staying away from the lamp''). VLMs could be used to score images of diverse robot paths obtained by robot motion planning algorithms, so as to select the path that best describes the user's preference. However, we still need to better understand how good current VLMs are at this task.

In this paper we generate a dataset of images of diverse robot motions in home settings, and evaluate the capability of VLMs to select the images that correctly match a text description of the displayed robot motion. To integrate VLMs into motion planning pipelines, we propose to proceed as follows: given a motion planning task (start and goal position of the robot), we generate diverse robot trajectories that achieve the task using a sampling-based motion planner, then we use a VLM to score the trajectories based on how well they match the user's description of spatial object relations or path styles, and finally we select the highest-score trajectory as the output.

Our work allows to evaluate the accuracy of different VLMs and image querying approaches at selecting these highest-score trajectories, thus informing future human-centered robot motion planning pipelines.

\section{RELATED WORK}

The visual understanding capability of VLMs has been applied to various tasks such as visual question answering~\citep{llava-liu2024improvedbaselinesvisualinstruction,spatialVLM_Chen_2024_CVPR}, object detection~\citep{chen2024taskclipextendlargevisionlanguage,kuo2023fvlmopenvocabularyobjectdetection} and optical character recognition~\citep{nacson2024docvlmmakevlmefficient}.
In robotics, VLMs and Large-Language Models (LLMs) have shown potential in enhancing the generalization capability of robots~\citep{brohan2023rt2visionlanguageactionmodelstransfer,shah2022lmnavroboticnavigationlarge,saycan2022arxiv,driess2023palme,bucker2024groundingrobotpoliciesvisuomotor,black2024pi0visionlanguageactionflowmodel} and assisting human-robot interactions~\citep{bucker2023latte,arkin2020multimodal}, even though they have also shown to be prone to various issues of bias~\citep{azeem2024llm,hundt2022robots}.

Various studies have used VLMs in robot task planning. For example, SayCan~\citep{saycan2022arxiv} leverages the semantic knowledge of LLMs to estimate the probability that actions make progress towards a given task. PaLM-E~\citep{driess2023palme} has demonstrated that large multimodal embodied models trained on robotics tasks and internet-scale data can address various tasks in the real world.
PaLM-E can generalize zero-shot on tasks with unseen objects, while SayCan is evaluated in kitchen tasks in a zero-shot manner.
Our method also uses VLMs zero-shot on robot tasks in home settings, but focuses on motion planning tasks. 
Various other methods have been proposed since, also for task planning~\citep{Honerkamp_2024,wang2022languagemodelsimagedescriptors,chen2022openvocabularyqueryablescenerepresentations,chen2023leveraginglargevisuallanguage,LinAgiaEtAl2023}.
LLMs can provide the spatial relationships between objects for task and motion planners in object rearrangement tasks~\citep{ding2023taskmotionplanninglarge}, which a planner can in turn use to compute task and motion plans.
Wang el al. generates natural language narration using multimodal inputs including environment observations, robot states and task planner~\citep{wang2024itellidoing}. The natural language grounding is shown to improve recovery efficiency.
NARRATE~\citep{ismail2024narrate} uses language to generate sequences of sub-tasks formulated with safe constraints and combine them with model predictive control.
Similarly, we use VLMs to process requirements specified in textual descriptions, but focus on generating robot trajectories from instructions.
Some methods use LLMs to break the long-horizon and complex tasks into sub-tasks and sub-goals \citep{yang2024guidinglonghorizontaskmotion,LinAgiaEtAl2023}. A recent study uses the reasoning capability of VLMs on raw images in trajectory planning~\citep{tang2025vlmplanner}. 
The previous works focus on using language models to compute goals and subgoals, but do not consider language to describe the style or desired properties of the path itself. 

Language models have shown potential to generate motion from textual instructions as well. MotionGPT~\citep{jiang2024motiongpt} generates consecutive human motions by using multimodal signals as input tokens in LLMs.
Researchers have also used Generative Pre-trained Transformer (GPT) with discrete representations to generate human motions from textual instructions~\citep{zhang2023t2mgptgeneratinghumanmotion}. Compared to these works, in this paper we use VLMs to generate robot motion instead of human motion, and our focus is on the use of VLMs (i.e. using images judges of proposed motion candidates).

Finally, another type of work in the literature using VLMs for robot planning is IMPACT~\citep{ling2025impactintelligentmotionplanning}, which uses these models to identify objects that tolerate contact with robots. While the approach is similar to our work in spirit, our work targets another problem---of modulating motion style or constraints using text input.

\section{METHOD}

\subsection{Generating Diverse Path Candidates with Motion Planning}

To evaluate the VLMs on robot planning tasks, we first use a heuristic method to obtain a diverse set of candidate paths and visualize them on an image of the robot and environment. In particular, we use a combination of the Bidirectional Rapidly-exploring Random Trees (BiRRT) \citep{Qureshi_2015,karaman2011samplingbasedalgorithmsoptimalmotion} and Probabilistic RoadMaps\citep{prm} to obtain diverse paths that satisfy start-state and goal-state constraints, though other diverse-planning methods could be used.

BiRRT operates by growing two trees: one from the start and one from the goal. It alternates between extending these trees and connecting them when they are sufficiently close. In order to obtain a variety of path candidates with different geometric and topological properties, we run BiRRT $n$ times using different random seeds, therefore leading to $n$ candidate trajectories.
Similarly, we run PRM $n$ times using different random seeds, each time building a graph from scratch and running Dijkstra to find a path to the goal, and each time using a different cost function for extra diversity: 
we used shortest distance, sinusoidal, and circular path costs.

Then, we use the K-means clustering algorithm to group trajectory candidates based on their waypoints, and partition the $n$ candidates into $k$ clusters. 
After clustering, the path that is closest to the centroid of each cluster is selected to be visualized in image inputs for VLMs, therefore leading to $k<n$ candidate paths (and $k$ images representing the execution or overlay of those paths in the environment, as we will describe next).

\subsection{Dataset construction}

Using the above method for obtaining diverse trajectories for a given motion planning problem, we then constructed a dataset of 558 language-constrained robot motion planning problems. Out of these, 126 were navigation problems, and 432 were arm-manipulation problems. Each problem consists of a virtual scene, a start and a goal location, and a text-description of the desired properties of the motion. We manually selected start and goal locations in the different scenes so as to allow for diverse ways of traveling between the two. More information about the scenes is provided in Section~\ref{sec:experimental-setup}.

We consider the following two categories of user preferences: 
\begin{itemize}
    \item \textbf{Object proximity} preferences describe desired spatial relationships with objects in the environment, e.g. stay close to/ away from object A, pass between object B and C.
    \item \textbf{Path style} preferences describe desired stylistic and shape-related properties of the robot trajectories, e.g. straight line, curved, zigzag, shortest path.
\end{itemize}

\begin{figure*}[tbp]
    \centering
    \includegraphics[width=0.8\textwidth]{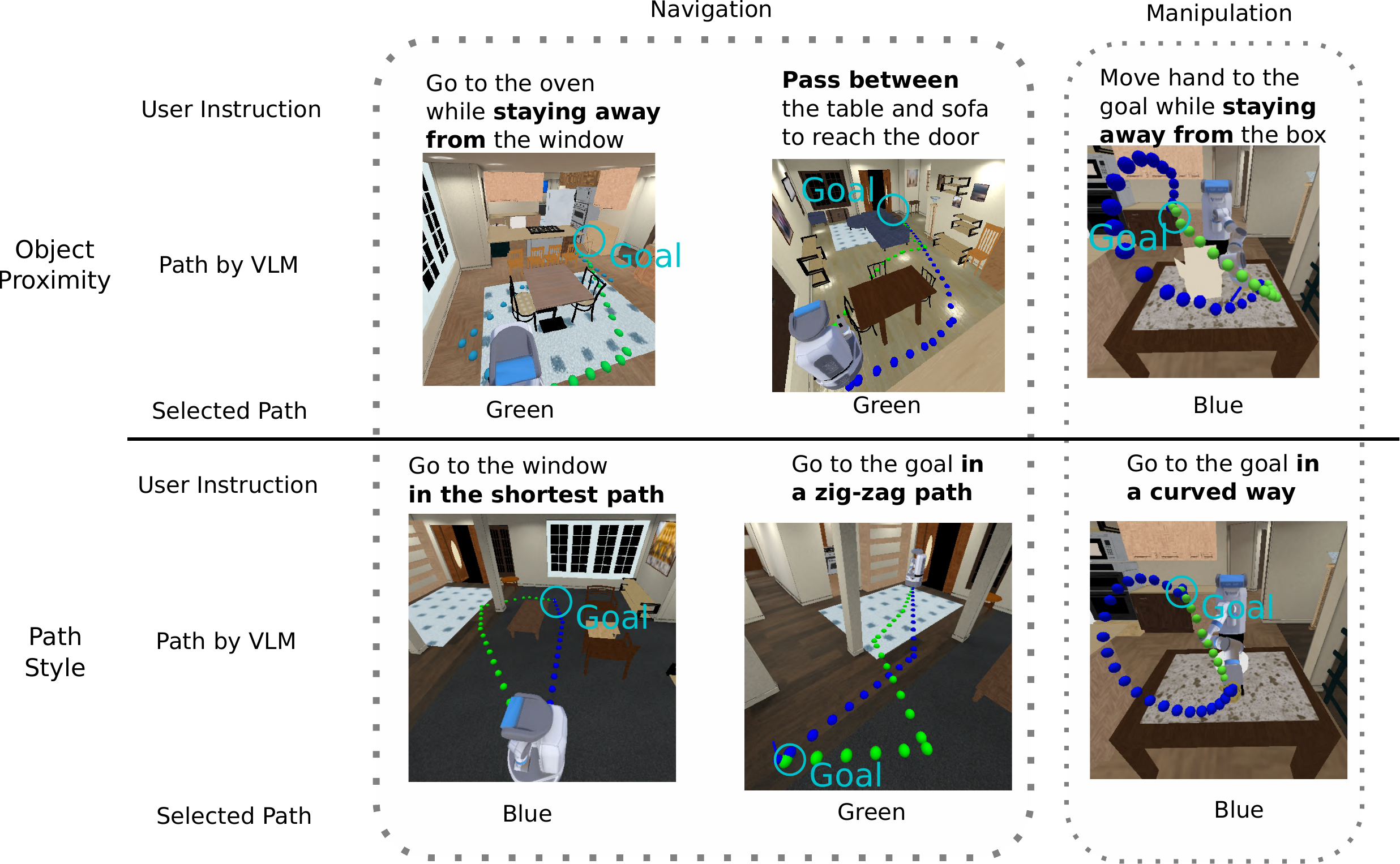}
    \caption{Examples of language-constrained robot motion planning problems, and the solutions scored highest by Qwen2.5-VL (dotted trajectories).}
    \label{fig:examples-problems-and-solution}
\end{figure*}

Examples are shown in Fig.~\ref{fig:examples-problems-and-solution}.
In order to obtain the ground-truth of the path that should be selected by a VLM for each problem, we applied our path-candidate generation and clustering procedure to generate a set of path candidates for each problem, and manually annotated each path with a description that was unique to it and could distinguish it from the others. In this way, each instruction is paired with a ground-truth path identifier. The curated dataset consists of sets of (image with multiple paths, text instruction, ground-truth path identifier for instruction).
Using this dataset, we then evaluated the accuracy of the VLM at selecting the right path candidate for each problem. 

\subsection{VLM-based Path Selection}

Given the candidate paths which are output by the clustering algorithm, we generate an image (or set of images) presenting those paths, and then use a Vision Language Model (VLM) to compute scores according to how well the image matches the instruction $l$, and obtain the path with the highest score.
We use the following prompt in order to obtain the selected path: ``This picture shows a robot in a simulated environment. Each trajectory of dots is a sequence of waypoints that guide the robot to move along. Please rate each of the trajectories out of 100 for how well they match the user instruction, by analyzing if they satisfy the constraints on object proximity and path style. User instruction: \{$l$\}. Give the color with the highest score at the end.''

\begin{figure}[t]
    \centering
    \subfloat[\centering ]{
        \begin{tikzpicture}
            \node[anchor=south west,inner sep=0] (image) at (0,0) {\includegraphics[width=0.19\columnwidth]{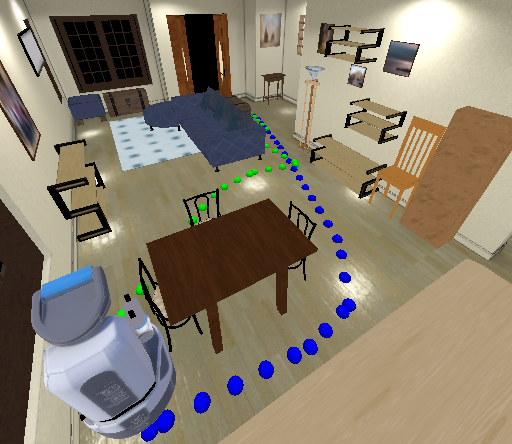}};
            
        \end{tikzpicture}
    }
    \qquad
        \subfloat[\centering ]{
        \begin{tikzpicture}
            \node[anchor=south west,inner sep=0] (image) at (0,0) {\includegraphics[width=0.19\columnwidth]{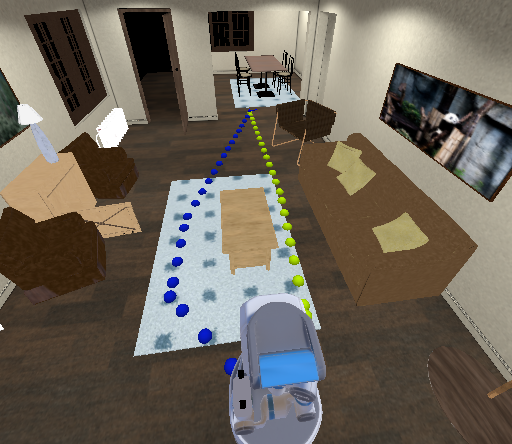}};
        \end{tikzpicture}
    }
    \qquad
    \subfloat[\centering ]{
        \begin{tikzpicture}
            \node[anchor=south west,inner sep=0] (image) at (0,0) {\includegraphics[width=0.18\columnwidth]{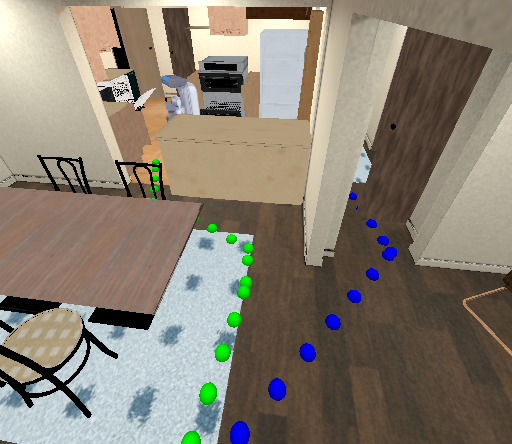}};
        \end{tikzpicture}
    }
    \qquad
    \subfloat[\centering ]{
        \begin{tikzpicture}
            \node[anchor=south west,inner sep=0] (image) at (0,0) {\includegraphics[width=0.18\columnwidth]{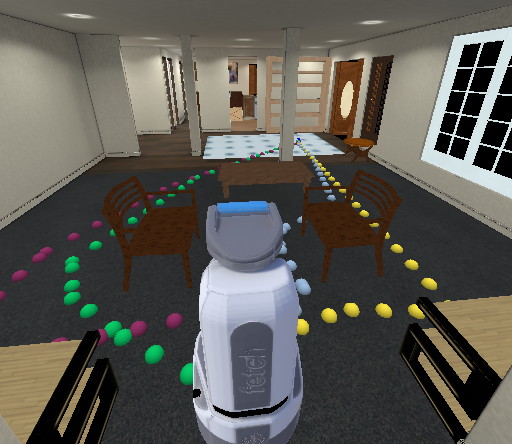}};
        \end{tikzpicture}
    }
    \caption{Examples of single-image trajectory trails in 4 scenes: a) move towards the door; b) move to the table in the kitchen; c) move to the table; d) move to the shelf behind the columns.}%
    \label{fig:example_single-image_traj}
    \vspace{-0.5em}
\end{figure}

\begin{figure}[t]
    \centering
    \subfloat[\centering ]{
        \begin{tikzpicture}
            \node[anchor=south west,inner sep=0] (image) at (0,0) {\includegraphics[width=0.45\columnwidth]{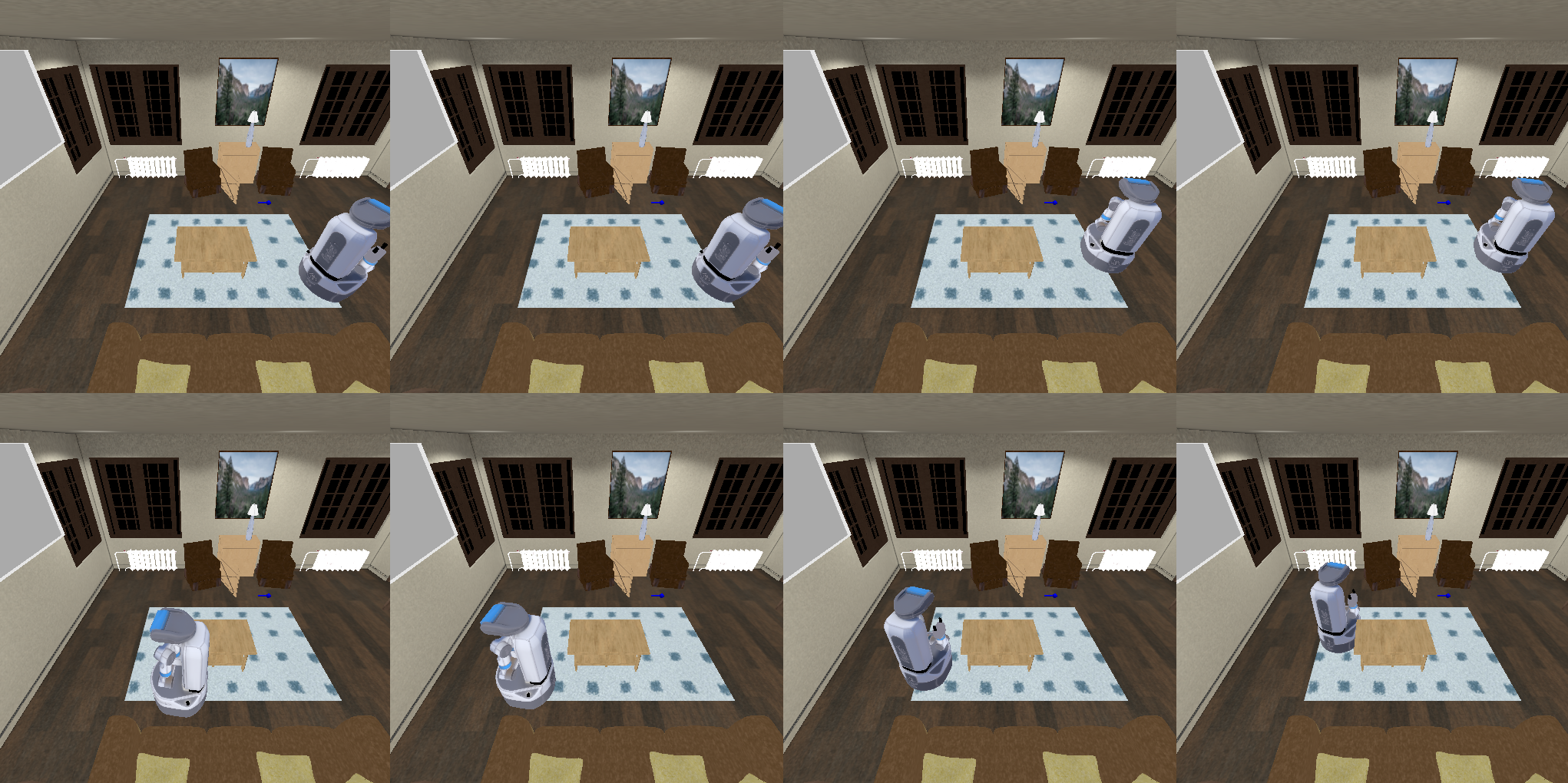}};
            
        \end{tikzpicture}
    }
    \qquad
        \subfloat[\centering ]{
        \begin{tikzpicture}
            \node[anchor=south west,inner sep=0] (image) at (0,0) {\includegraphics[width=0.45\columnwidth]{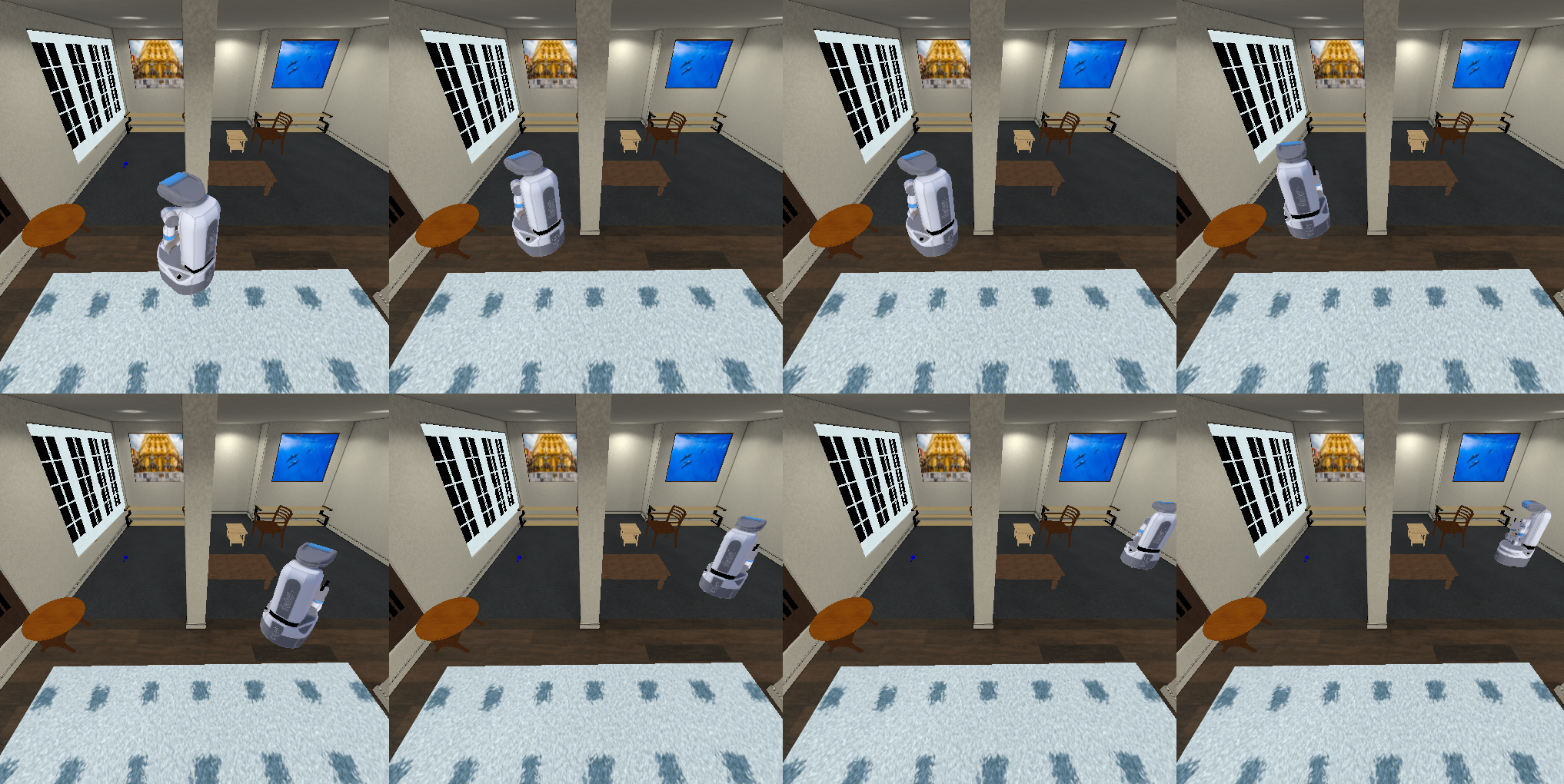}};
        \end{tikzpicture}
    }
    \caption{Example of galleries of screenshots showing the robot moving along each trajectory, with each row presenting a sequence screenshots from one trajectory. (a) The robot goes around the table in 2 topologically different ways; (b) The robot goes around the table and column in 2 topologically different ways.}
    \label{fig:example_scr_gallery}
\end{figure}

We propose four methods of querying VLMs for robot path selection from images:
\begin{enumerate}
    \item Single-image trajectory: the candidate paths are shown as trails of dots in different colors, as shown in Fig. \ref{fig:example_single-image_traj}. For navigation problems, we visualize paths as dots along the route of the base of the robot. For manipulation problems, we visualize the trajectory of the end-effector with similar dots (e.g. Fig. \ref{fig:examples-problems-and-solution}). All the candidate paths are shown in one picture to query VLMs. 
    \item Multi-image trajectory trail: all the candidate paths are split into individual images, where each image contains one trajectory trail. The VLM is queried once for every image, to give scores to each image based on the instructions. Finally we select the image (trajectory) with the highest score.
    \item Single-image trajectory trail with visual context: the VLM is first asked to generate visual context on the images \citep{maeda2024vision_visual-context}, which describes the detailed content of images with structured data, including objects, features, and relationships between them. Then the visual context is given to the VLM together with textual and visual prompts to output the highest-score path.
    \item Single-image screenshot gallery: for each candidate path we generate a sequence of images of the robot moving along the path by simulating the execution of the path. Fig. \ref{fig:example_scr_gallery} shows the screenshot gallery of 2 candidate trajectories, with each row representing one path. The VLM is then asked to select the row that satisfied the user description.
\end{enumerate}

We will evaluate and compare these four methods in our experiments.

\section{RESULTS}

\subsection{Experimental setup}
\label{sec:experimental-setup}

We used iGibson~\citep{igibson}, a simulation environment that contains 3D interactive scenes reconstructed from real homes, to generate robot trajectories in simulated home environments. As described in the previous section, we used the BiRRT motion planner~\citep{Qureshi_2015,karaman2011samplingbasedalgorithmsoptimalmotion} and Probabilistic RoadMap\citep{prm} to generate $n=$50 diverse paths from the start to goal in each planning problem in each simulated scene. We use a set of iGibson scenes called Ihlen\textunderscore 1\textunderscore int, Pomaria\textunderscore 1\textunderscore int, Beechwood\textunderscore 0\textunderscore int, Benevolence\textunderscore 1\textunderscore int and Merom\textunderscore 0\textunderscore int which include objects from the BEHAVIOR dataset~\citep{srivastava2022behavior}.
We evaluate three different VLMs: 
1) Qwen2.5-VL\citep{qwen2.5-VL}\citep{Qwen2VL}: a VLM with visual localization ability. It uses bounding boxes and point-based representation for object grounding which provides a foundation for visual reasoning. We use the version Qwen2.5-VL-72B-Instruct in our experiments.
2) GPT-4o: a model that can take images as input and answer questions about the images. It achieves state-of-the-art performance on visual perception benchmarks including MMMU~\citep{yue2024mmmumassivemultidisciplinemultimodal} and ActivityNet~\citep{activitynet_benchmark}. 3) LLaVa1.5: a model that excels in conversational-style Visual Question Answering~\citep{liu2023llava}.

\subsection{Qualitative analysis}

Fig.~\ref{fig:examples-problems-and-solution} shows 6 example problems with user instructions and the motion scored highest by Qwen2.5-VL.
The figure shows that the robot is instructed to go to the goal position with constraints on the movement with respect to certain objects in the environment or the overall path style. 
In the first example (top left), the robot is commanded to go to the oven. There are 2 ways to go around the table to reach the goal position. The VLM selected the trajectory that matches the instruction which is far away from the window. 
In the last example (bottom right), the robot receives the instruction to move its hand in a curved way. In this case, the blue curved path is chosen instead of the green straight line.

\subsection{Accuracy with different image query methods}

\begin{figure}[t]
    \centering
    \includegraphics[width=0.6\columnwidth]{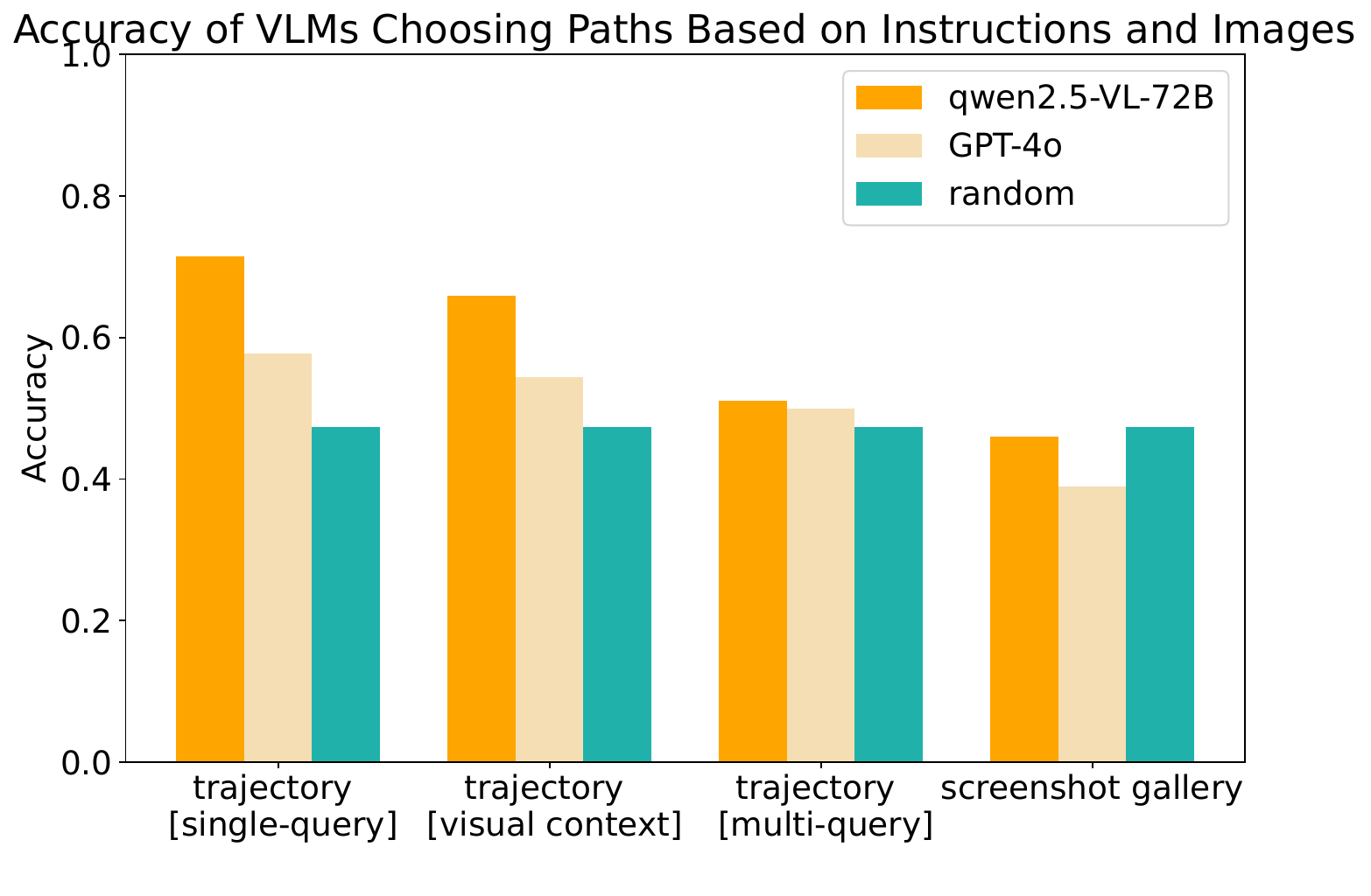}
    \caption{Accuracy of VLMs in selecting candidate paths in images with different query methods, averaged across two types of preferences (object proximity and path style).}
    \label{fig:plot_succ_rate}
\end{figure}

Fig. \ref{fig:plot_succ_rate} shows overall accuracy on the navigation problems, using the 4 different image query methods. 
As the figure shows, the best-performing method was the single-query method, with an average accuracy over 70\% when using Qwen2.5-VL.
The multi-query method led to lower accuracy, even though it uses more queries.
We hypothesize that the reason for this is that when a VLM is asked to score a trajectory individually, the metric is not consistent across multiple requests---because it is not able to compare the one in the image currently being processed with the others.
On the other hand, when all the trajectories are shown together in the single-query method, the VLMs can compare which trajectory is further away from the painting and assign consistent relative scores to those trajectories. 

The screenshot-gallery method led to an accuracy slightly higher than random selection. Compiling all screenshots in one image makes the size of each screenshot small, thus leading to a lack of detail. 
For the rest of the experiments, we therefore use the single-query method, which led to highest accuracy.

\subsection{Accuracy on two types of preferences (object proximity and path style)}

\subsubsection{Navigation Tasks}
\begin{figure}[t]
    \centering
    \includegraphics[width=0.6\columnwidth]{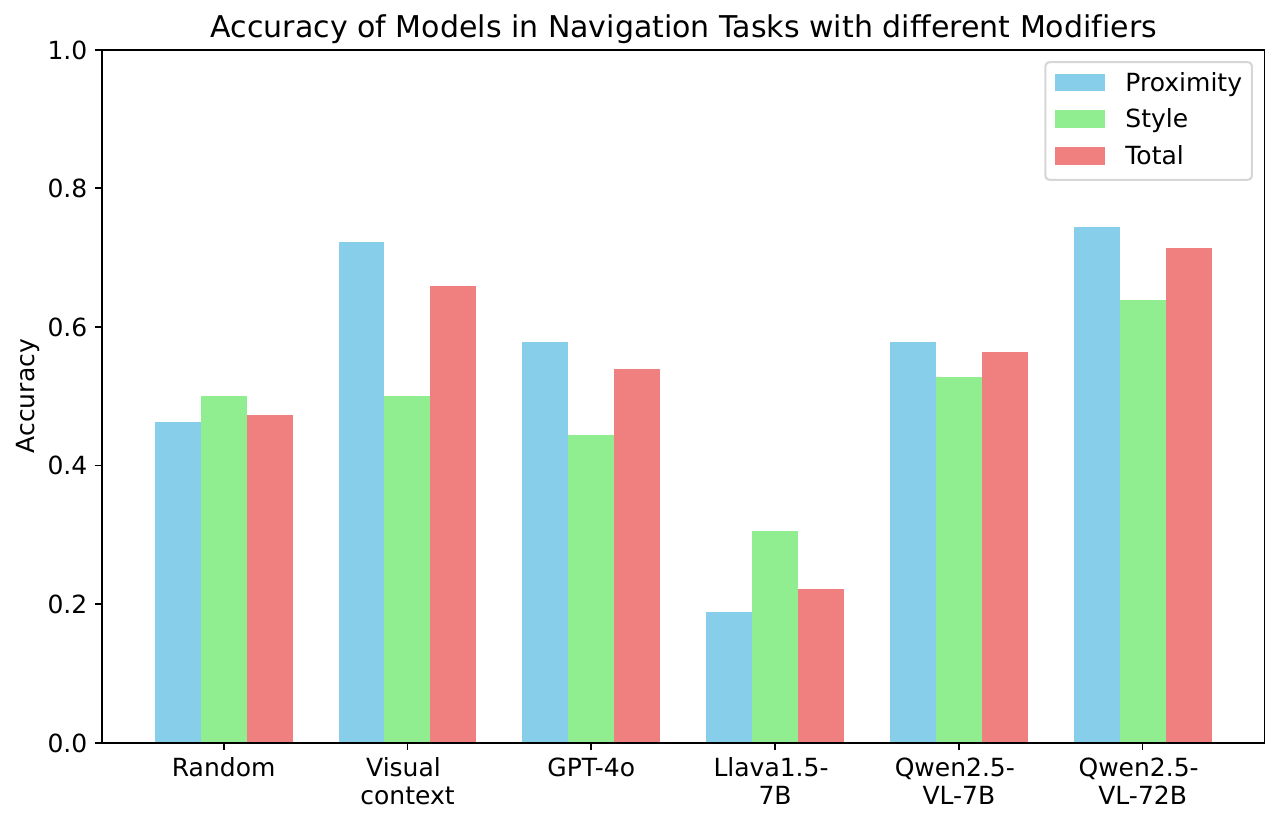}
    \caption{Accuracy of VLMs in selecting candidate paths in images, on navigation tasks.}
    \label{fig:plot_succ_rate_modifier}
\end{figure}

The accuracy on different types of preferences on the 126 instruction-image pairs (90 object-proximity and 36 path-style) is shown in Fig.~\ref{fig:plot_succ_rate_modifier}. All methods are evaluated zero-shot. Qwen2.5-VL performs better than GPT4-o, which is consistent with its reported precise object grounding capabilities~\citep{qwen2.5-VL}. With single-query trajectory trail image, Qwen2.5-VL achieves an accuracy of 74.4\% on proximity problems and 63.9\% on path-style problems. For all the tested models except Llava1.5, the accuracy of proximity problems is higher than path style problems.

From our experiment, adding a visual context step does not increase accuracy. This might be due to the fact that these large-sized models automatically generate and keep tokens of image context already, e.g. Qwen2.5 has a default setting to keep context with length up to 32,768 tokens. Although visual context can help smaller-sized models such as Llava~\citep{llava-liu2024improvedbaselinesvisualinstruction} to keep track of detailed content of images, including objects, attributes, and relationships~\citep{maeda2024vision_visual-context}, this extra step can add redundancy to the textual information and can potentially undermine the performance of models built with internal context tracking.

\subsubsection{Manipulation Tasks}
\begin{figure}[t]
    \centering
    \includegraphics[width=0.6\columnwidth]{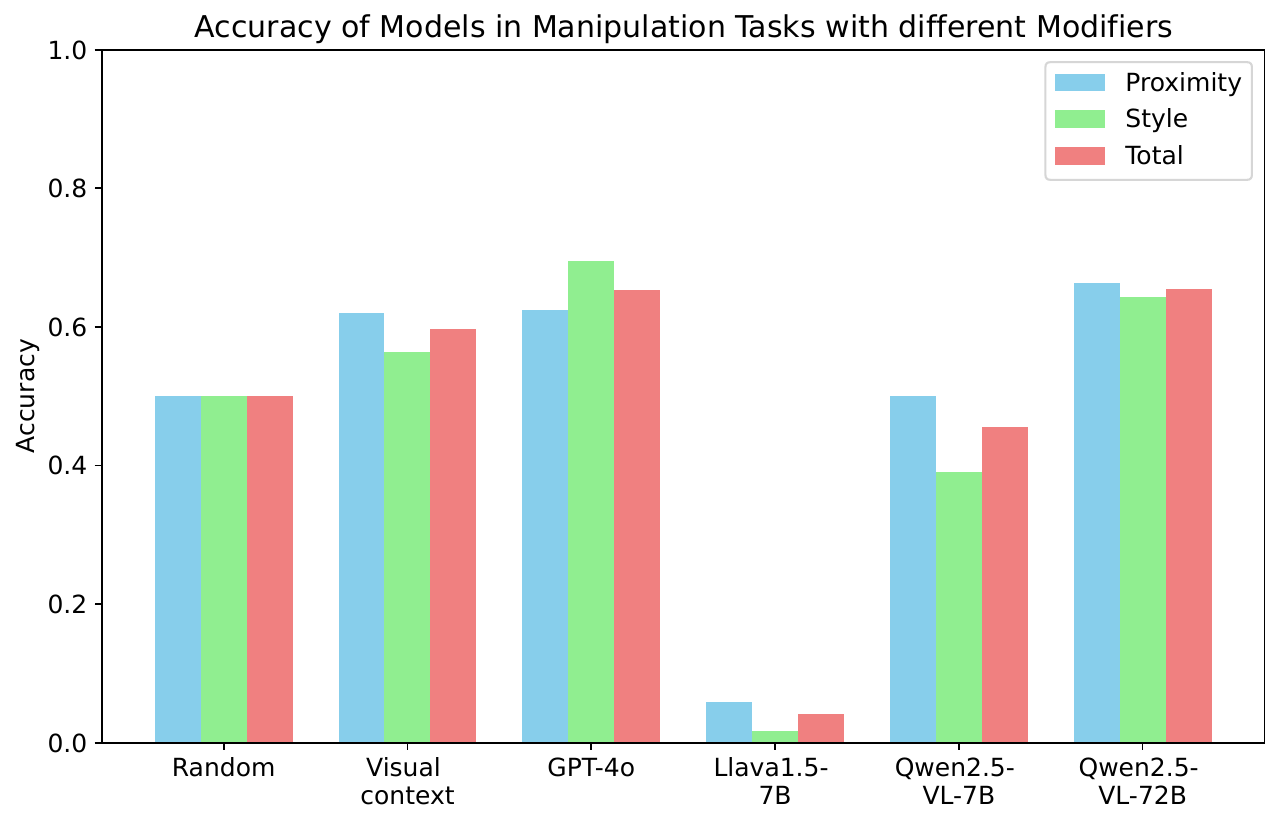}
    \caption{Accuracy of VLMs in selecting candidate paths in images, on manipulation tasks.}
    \label{fig:plot_manip_succ_rate_w_modifiers}
\end{figure}

The accuracy of the visually matching the textual command to robot taks image on 432 manipulation problems (258 proximity and 174 path style) is shown in Fig. \ref{fig:plot_manip_succ_rate_w_modifiers}. In the manipulation tasks, the robot stands in front of a table with objects and is asked to move its arm in reference to the objects on the tabletop. Examples are shown in Fig. \ref{fig:examples-problems-and-solution}. Qwen2.5-VL-72B has the highest accuracy in proximity problems (66.3\%) and GPT-4o has the highest in path-style problems (69.5\%). The overall success rate in manipulation tasks is lower than in navigation (65.5\% in manipulation vs 71.4\% in navigation on Qwen2.5-VL-72B).

\subsection{Computation cost}

\begin{figure}[t]
    \centering
    \includegraphics[width=0.6\columnwidth]{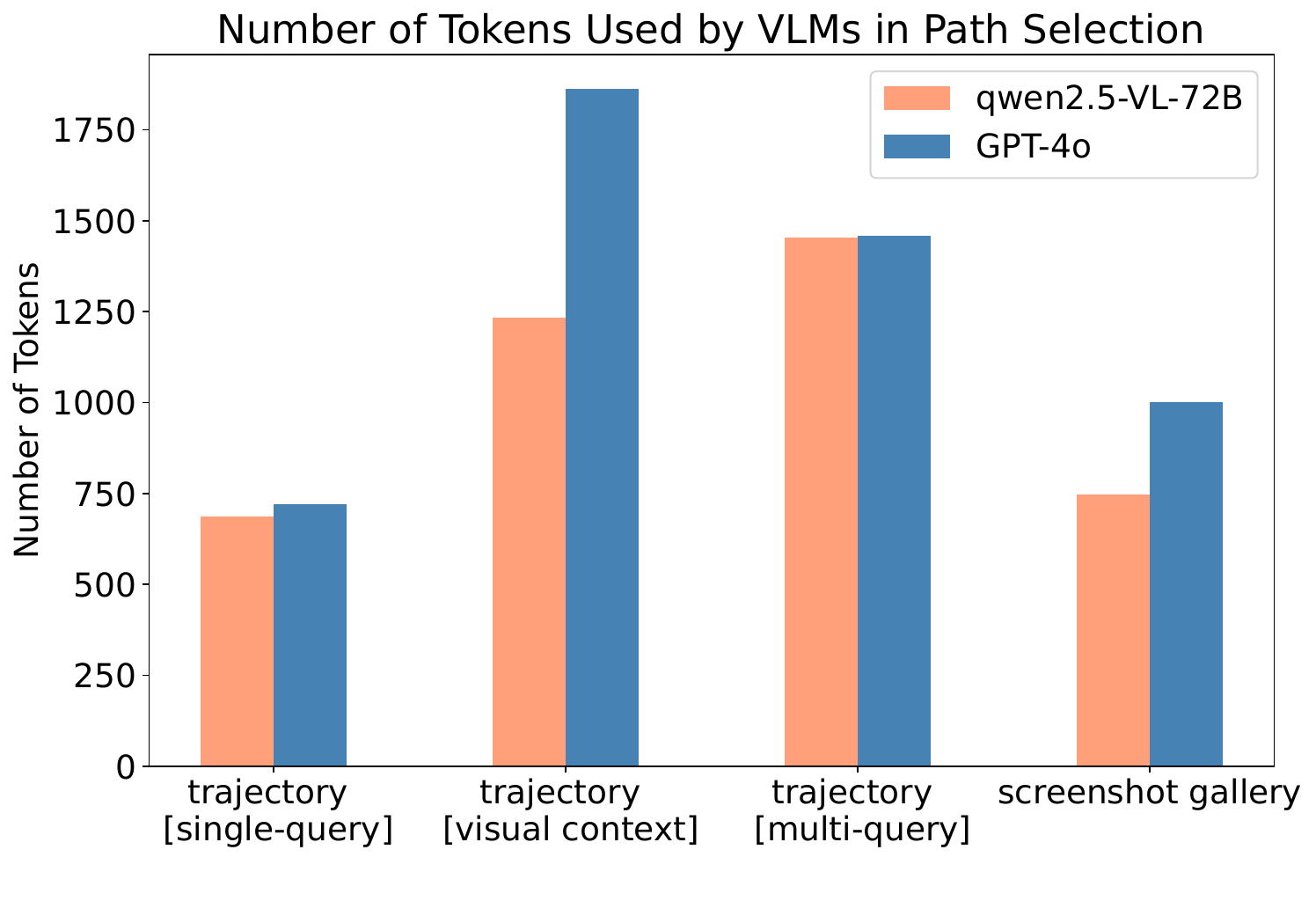}
    \caption{Number of tokens used by each querying method.}
    \label{fig:plot_tokens_used}
\end{figure}

We investigate the computation cost of each querying method by comparing the number of tokens used by each method and each VLM. Fig.~\ref{fig:plot_tokens_used} shows the average number of tokens used in each request to the VLMs. When all trajectories are shown in one image, an average token number of 687.3 is used by Qwen2.5-VL and 721.2 by GPT-4o. This querying method requires the fewest number of tokens among all methods we have tested, since it uses a single image, while also leading to the highest performance.

\begin{figure}[t]
    \centering
    \includegraphics[width=0.45\columnwidth]{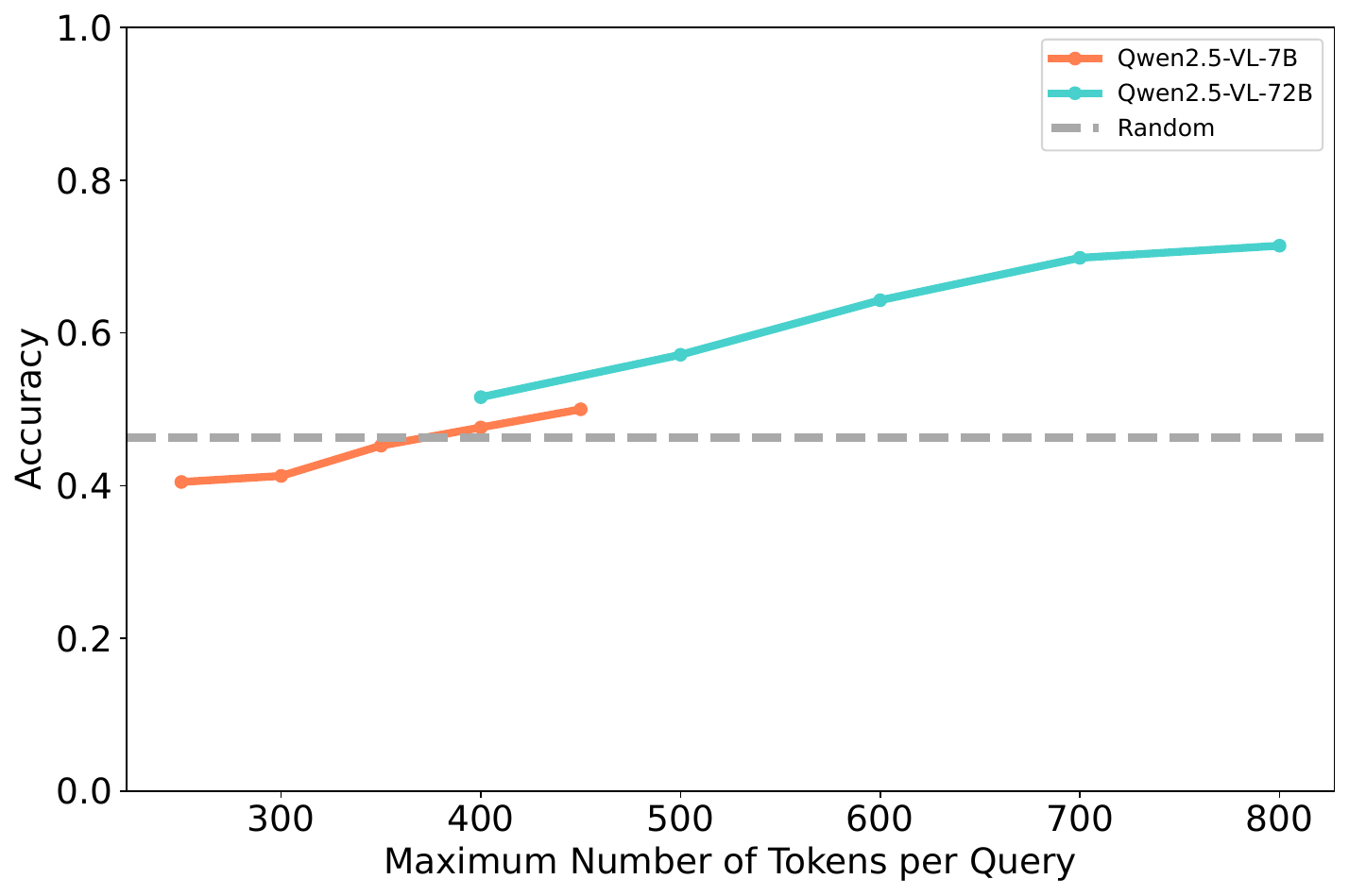}
    \caption{Accuracy of path selection with varying max. number of allowed tokens (image size).}
    \label{fig:acc-token}
\end{figure}

Next, we investigate the relation between the \textit{financially-relevant} computation cost, in number of tokens, that VLMs consume and the accuracy of the path selection by limiting the maximum number of tokens used (i.e. resizing the images sent to the VLMs). We test the accuracy of the models by restricting the maximum allowed number of token from 200 to 800 per query on Qwen2.5-VL-7B and 72B. Fig.\ref{fig:acc-token} shows that, given a fixed querying method (single query trajectory image), the accuracy of path selection grows approximately linearly with the number of tokens used within the range between the worst and best performance. 

\subsection{Effect of Finetuning on Language Models with smaller size}

\begin{figure*}[htbp]
\centering
\begin{subfigure}{0.31\textwidth}
    \centering
    \includegraphics[width=\linewidth]{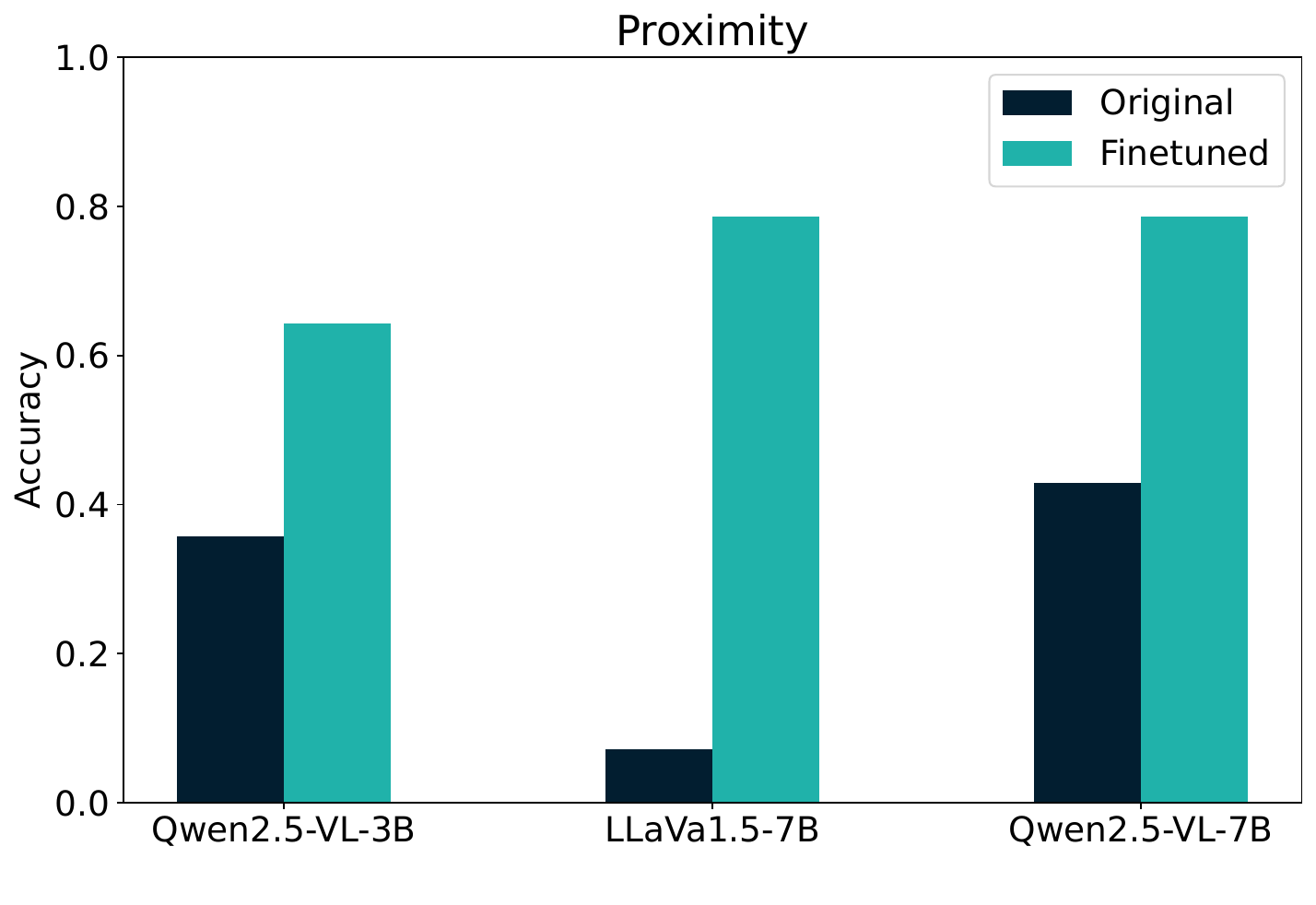}
    \caption{Proximity}
\end{subfigure}
\begin{subfigure}{0.31\textwidth}
    \centering
    \includegraphics[width=\linewidth]{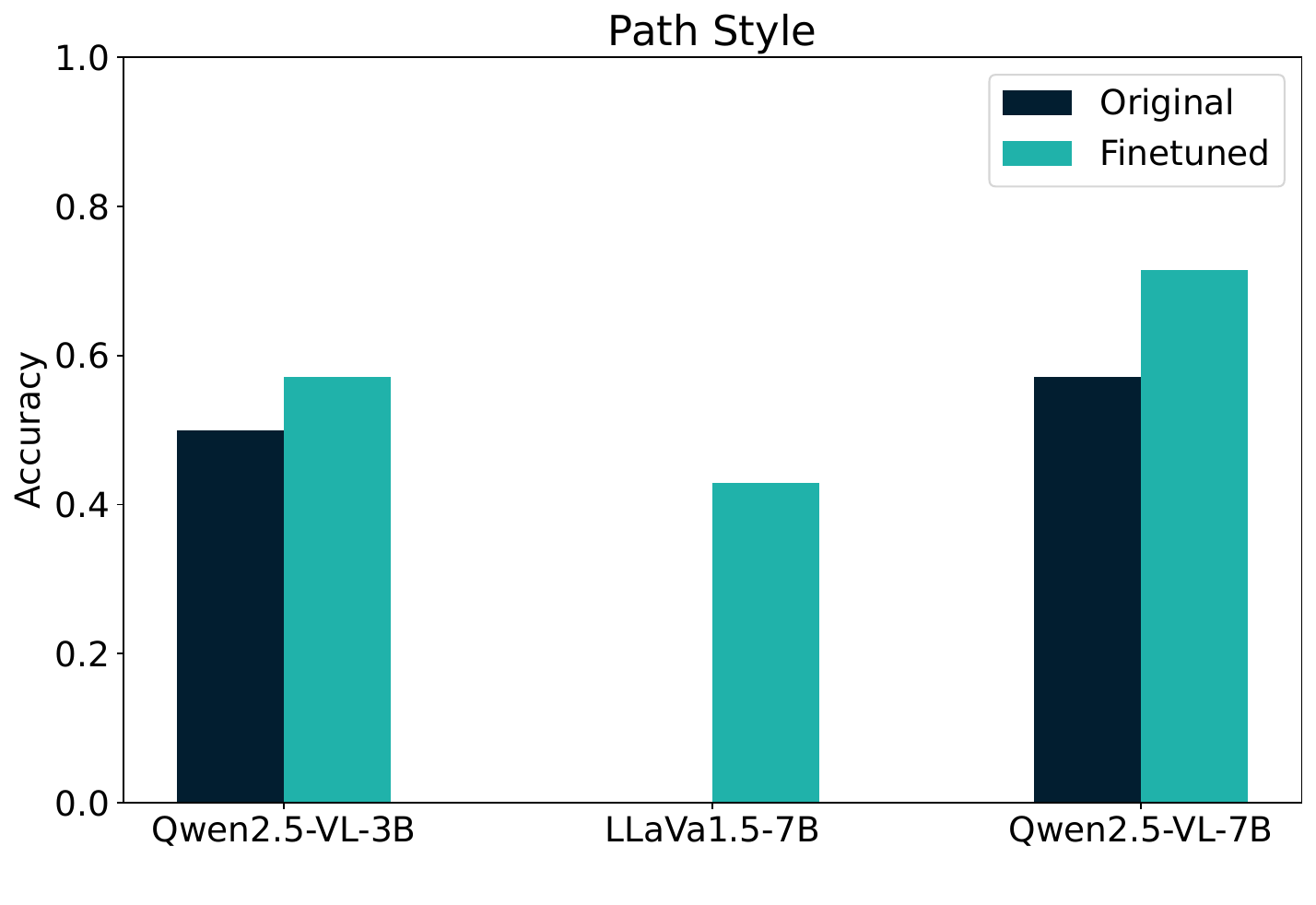}
    \caption{Path style}
\end{subfigure}
\begin{subfigure}{0.31\textwidth}
    \centering
    \includegraphics[width=\linewidth]{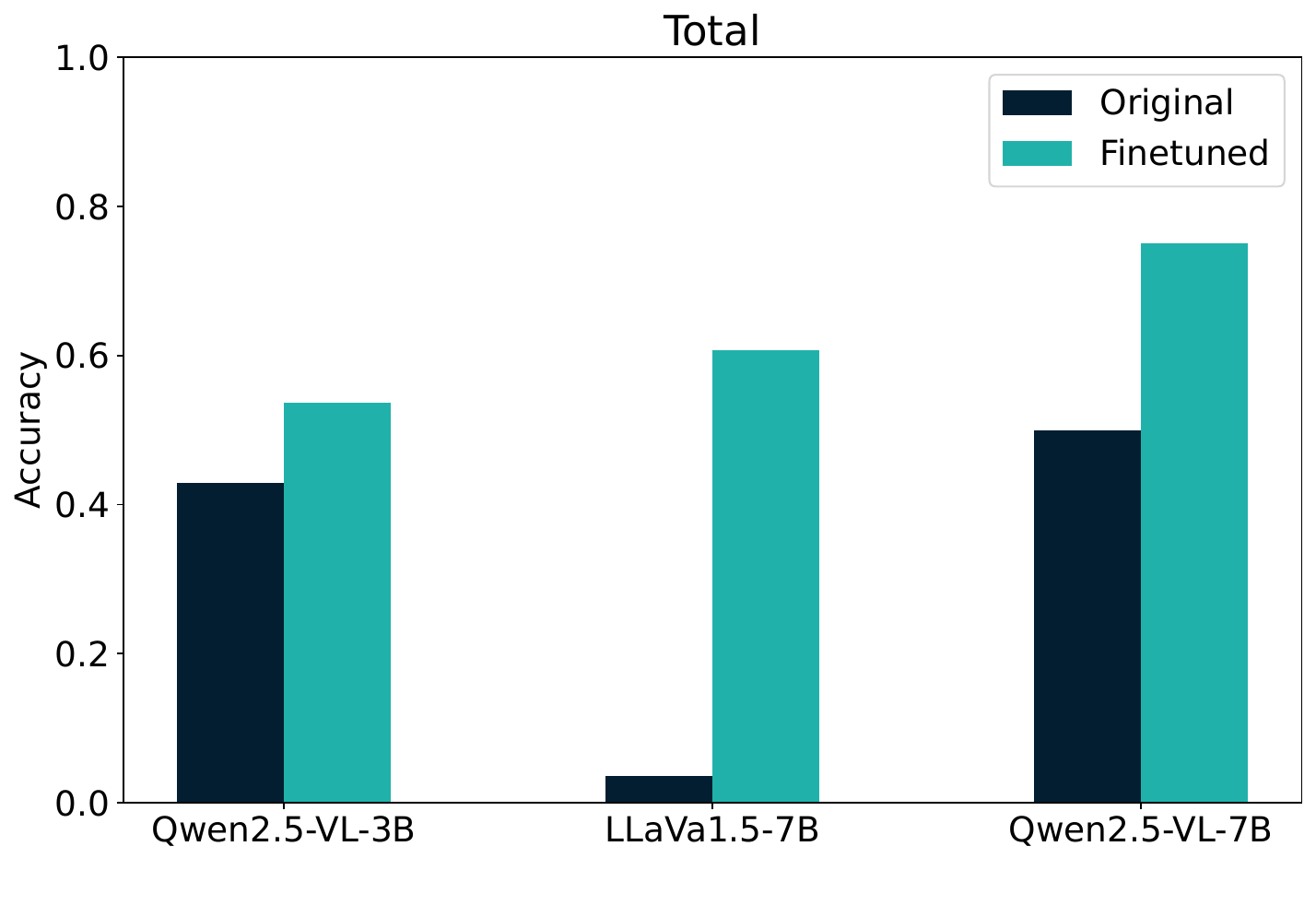}
    \caption{Total}
\end{subfigure}

\caption{Accuracy of zero-shot and finetuned models on a test set of 28 problems.}
\label{fig:finetuned}
\end{figure*}

Finally, we studied the performance improvement after finetuning. We used Supervied Fine-Tuning (SFT) with a training set that contained 98 examples on 2 models, LLaVa1.5-7B and Qwen2.5-VL-7B. We then tested them on a test set that contained 14 proximity problems and 14 path-style problems. The performance of the models before and after finetuning is shown in Fig. \ref{fig:finetuned}. 
For fair comparison purposes, values of zero-shot performance of the original models shown on the table were also computed on the test.
The figure shows there is a prominent increase in accuracy in proximity problems for all models after finetuning. The accuracy improves more than 20\% for Qwen2.5-VL-7B and more than 60\% for LLaVa1.5-7B after exposure to a small dataset. Therefore, the architecture can adapt to new types of user instructions. Although the smaller model (Qwen2.5-VL-3B) takes less computation for finetuning and inference, its performance did not improve as much as other 2 models with 7B parameters.

\subsection{Discussion and Limitations}

As shown in Fig.~\ref{fig:plot_succ_rate}, using different methods of visualization to query VLMs can lead to varied performance. We found that a single image showing all clustered candidate trajectories as colored dotted paths was an effective way to query state-of-the-art VLMs to select robot motion, given user descriptions of motion preferences. Different visualization methods could lead to further improvements.

We found that a common failure was that VLMs failed to recognize which path was the shortest or longest when requested, which is exactly the type of problem classical optimal planners (e.g. RRT*, PRM*) can efficiently solve. Another common failure was of ``hallucination'', where VLMs selected a candidate path that did not exist (e.g. selected `red' as the best path, when none were of that color).

\section{CONCLUSIONS}

In this paper, we evaluated the spatial reasoning capabilities of VLMs to select robot trajectories that satisfy user preferences related to path topology and style.  
VLMs scored motion candidates with respect to how well they matched the user request, given images with candidate paths generated by robot motion planners. We showed that Qwen2.5-VL achieves an overall accuracy of 71.4\%, with 74.4\% on proximity and 63,9\% on path style problems, which is considerably higher than GPT4-o. The evaluated VLMs performed better on object-proximity preferences, compared to path style (e.g. to obtain a ``long'' or ``zigzag'' path), but finetuning was able to obtain high accuracy gains even on small vision-language models with few examples, by between 10 and 57 percentage points.
Finally, we investigated the tradeoff between accuracy and computation cost, showing that accuracy decreases approximately linearly with the decrease in number of tokens or image size.
In the future, our goal is to integrate these capabilities robustly into robot motion planning pipelines, though this may require significant work on increasing accuracy or integrating users into the loop through appropriate interfaces.

\bibliography{iclr2026_conference}
\bibliographystyle{iclr2026_conference}


\end{document}